\documentclass[review]{elsarticle}

\usepackage{lineno,hyperref}
\modulolinenumbers[5]

\usepackage{graphicx}
\usepackage{comment}
\usepackage{amsmath,amssymb} 
\usepackage{color}

\usepackage{url}
\usepackage{bm}
\usepackage{amsmath}
\usepackage{algorithm}
\usepackage{algorithmic}
\usepackage{amssymb}
\usepackage{booktabs}
\usepackage{subfigure}
\usepackage{textcomp}
\usepackage{multirow}
\usepackage{wrapfig}
\usepackage{caption}
\usepackage{comment}

\newcommand{\eg}{{\emph{e.g.}}}

\newcommand{\norm}[1]{\left\lVert#1\right\rVert}

\newcommand{\qileft}{[\kern-0.15em[}
\newcommand{\qiLeft}{\left[\kern-0.4em\left[}
\newcommand{\qiright}{]\kern-0.15em]}
\newcommand{\qiRight}{\right]\kern-0.4em\right]}

\newcommand{\st}{{\mbox{s.t.}}}

\newcommand{\ie}{{\emph{i.e.}}}
\newcommand{\wrt}{{{w.r.t.}}}

\newcommand{\red}[1]{{\color{red}{#1}}}
\newcommand{\blue}[1]{{\color{blue}{#1}}}
\newcommand{\FLOPs}{\mbox{FLOPs}}
\newcommand{\FC}{\mbox{FC}}
\newcommand{\Sigmoid}{\mbox{Sigmoid}}
\newcommand{\ReLU}{\mbox{ReLU}}

\newcommand{\g}{{\bm{g}}}

\newcommand{\x}{{\bm{x}}}

\newcommand{\R}{\mathcal{R}}

\newcommand{\A}{{\mathcal{A}}}
\newcommand{\B}{\mathcal{B}}

\newcommand{\D}{\mathcal{D}}
\newcommand{\F}{\mathcal{F}}
\newcommand{\G}{\mathcal{G}}

\newcommand{\I}{\mathcal{I}}
\renewcommand{\L}{\mathcal{L}}
\newcommand{\M}{\mathcal{M}}
\newcommand{\N}{\mathcal{N}}

\newcommand{\Z}{\mathcal{Z}}
\newcommand{\W}{\mathcal{W}}
\renewcommand{\red}[1]{{\color{red}{#1}}}
\renewcommand{\blue}[1]{{\color{blue}{#1}}}

\journal{Journal of \LaTeX\ Templates}

\begin{document}

\begin{frontmatter}

\title{Data Agnostic Filter Gating for Efficient Deep Networks}


\author[mymainaddress]{Xiu Su\fnref{myfootnote}}
\fntext[myfootnote]{Work was done during internship at SenseTime.}
\ead[url]{xisu5992@uni.sydney.edu.au}

\author[mythirdaddress,mysecondaryaddress]{Shan You\corref{mycorrespondingauthor}}
\ead[url]{youshan@sensetime.com}

\author[mythirdaddress]{Tao Huang}
\ead[url]{huangtao@sensetime.com}

\author[myforthaddress]{Hongyan Xu}
\ead[url]{tjdxxhy@tju.edu.cn}

\author[mythirdaddress]{Fei Wang}
\ead[url]{wangfei@sensetime.com}

\author[mythirdaddress]{\\Chen Qian}
\ead[url]{qianchen@sensetime.com}

\author[mysecondaryaddress,myfifthaddress,mysixthaddress]{Changshui Zhang}
\ead[url]{zcs@mail.tsinghua.edu.cn}

\author[mymainaddress]{Chang Xu}
\ead[url]{c.xu@sydney.edu.au}

\cortext[mycorrespondingauthor]{Corresponding author}

\address[mymainaddress]{School of Computer Science, Faculty of Engineering, The University of Sydney}
\address[mythirdaddress]{SenseTime}
\address[mysecondaryaddress]{Department of Automation, Tsinghua University}
\address[myforthaddress]{School of Precision Instrument and Opto-Electronics Engineering, Tianjin University}
\address[myfifthaddress]{Institute for Artificial Intelligence, Tsinghua University (THUAI)}
\address[mysixthaddress]{Beijing National Research Center for Information Science and Technology (BNRist)}

\begin{abstract}
	To deploy a well-trained CNN model on low-end computation edge devices, it is usually supposed to compress or prune the model under certain computation budget (\eg, FLOPs). Current filter pruning methods mainly leverage feature maps to generate important scores for filters and prune those with smaller scores, which ignores the variance of input batches to the difference in sparse structure over filters. In this paper, we propose a data agnostic filter pruning method that uses an auxiliary network named Dagger module to induce pruning and takes pretrained weights as input to learn the importance of each filter. In addition, to help prune filters with certain FLOPs constraints, we leverage an explicit FLOPs-aware regularization to directly promote pruning filters toward target FLOPs. Extensive experimental results on CIFAR-10 and ImageNet datasets indicate our superiority to other state-of-the-art filter pruning methods. For example, our 50\% FLOPs ResNet-50 can achieve 76.1\% Top-1 accuracy on ImageNet dataset, surpassing many other filter pruning methods.
\end{abstract}


\end{frontmatter}


\section{Introduction}

Recently, artificial intelligence (AI) engines with deep learning techniques has achieved remarkable success in various tasks \cite{li2020speech,wang2018devil,du2020learnability,shi2019reinforced,wei2020point,ming2019group,yangdeep,yang2018shared}, and networks (\eg, convolutional neural networks, CNNs) are thus 
favored in the establishment of cloud and edge computing, which are mainly deployed in terminal devices, such as mobile phones, tablets, AR glasses, wearable watches, and onboard surveillance equipment. However, aiming at the state-of-the-art accuracy performance, conventional trained CNN models in the industrial model zoo usually have huge model size. And they are clumsy for deployment on low-end computational devices. In this way, a natural problem goes that besides the fundamental accuracy performance, how we can develop a ready-to-deploy model under certain computation budget, such as FLOPs. Luckily, due to the development of model compression and acceleration, pruning has been an efficient way to acquire light models based on existing clumsy models with foundation performance.

Currently, pruning can be divided into the categories of weight pruning or filter pruning. However, filter pruning is more competitive than weight pruning, since it can result to a lightweight model which has the consistent network structure of pre-trained model and is friendly to current off-the-shelf deep learning frameworks. Moreover, filter pruning is complementary to other compression techniques, the pruned networks can be usually further compressed using quantization \cite{ChenCompressing,han2015deep}, low-rank decomposition \cite{DentonExploiting,SindhwaniStructured} or knowledge distillation \cite{you2017learning,you2018learning,kong2020learning}. To prune the redundant filters, one effective way is to prune filters with pre-trained weights, and the filters with less importance to network performance are referred to as redundant filters in this paper. Basically, filter pruning works by first finding and pruning the redundant filters, then retraining (fine-tuning) the pruned network to recover its performance.

Identifying redundant filters matters in filter pruning. In specific, many filter pruning methods leverage scaling factors to find out the redundant filter, \eg~ scaling factors \cite{gate_decorator}, trainable auxiliary parameters \cite{autoprune}. However, all of these methods leverage filter-wise auxiliary parameters to determine the redundancy of filters, which is usually optimized simultaneously with network parameters in the form of multiplication, the collaboration or competition of these parameters may lead to unexpected result, \eg, the filter weights corresponding to the smaller auxiliary parameters may be large for balance, so it may inaccurate to judge the redundant filters directly from auxiliary parameters. Besides, to determine the redundancy of a specific filter in the overall convolution kernel, it is better to use the information of all feature maps or filters together rather than the filter-wise auxiliary parameters. To identify the filter redundancy globally, many filter pruning algorithms propose to construct a gate network by leveraging the feature maps as input to generate importance scores for filters. However, the variance of input batches may lead to the difference in sparse structure over filters. Thus those filters with non-support scores are to be pruned, where non-supporting\footnote{https://en.wikipedia.org/wiki/Support\_(mathematics)} gates represent a subset of gates mapped to zero values. However, since feature maps are subject to input batches, different batches may generate distinct importance scores and the corresponding support, which makes it difficult to determine an optimal score support for all input batches, and causes a performance gap accordingly. 

In addition, filter pruning methods mainly neglect the allocation of the FLOPs budget during training. In order to ensure the pruned network is under some FLOPs budgets, they have to resort to various sparsity proxies, \eg, $\ell_{21}$ norm of filters, and $\ell_{1}$ norm of filter weights or scaling factors. Nevertheless, the obtained network induced by these sparsity proxies is not necessarily optimal for the constrained FLOPs budget. And it usually needs a cautious hyper-parameter setup for the sparsity proxy so that the FLOPs of pruned network matches exactly with the given budget.

In this paper, we propose to prune the redundant filters through the Dagger module with kernel weights used as the input to deduce the redundancy of different filters, which has three folds of advantages. Firstly, gates of filters are generated based on the Dagger module, which avoids the joint optimization of gates and filter weights. Secondly, kernel weights are optimized through the whole training dataset for several epochs, which indicates the kernel weights contains information related to the whole dataset. Thirdly, we can easily and quickly adapt the pre-trained network to different budgets of pruned networks. Besides, based on the Dagger module, we also propose a FLOPs aware regularizer to directly pruning redundant filters from the pre-trained model with target FLOPs budget. Concretely, we allocate a binary gate for each filter where 0 means that the filter should be pruned and 1 is otherwise. In this way, the status of all binary gates corresponds to a certain filter configuration and FLOPs value. However, binary gates are hard to optimize, thus we relax the binary gates into real gates of the interval $[0,1]$, and model them by a Dagger module using the pre-trained weights. Therefore, these Dagger modules can sufficiently exploit the information within pre-trained filters, and serve as a decent surrogate for binary gates, thereby deriving a corresponding FLOPs regularization term. With the weights of the pre-trained model fixed, the gates generated by Dagger module can be optimized by maintaining the accuracy performance as well as reducing the FLOPs such that a smaller network can be induced. Besides, for cohering with a fixed FLOPs budget, we propose to optimize the Dagger module in a greedy manner, so that the model is pruned gradually and we can check whether the valid FLOPs of current pruned network matches with the budget.

We have conducted extensive experiments on benchmark CIFAR-10 \cite{krizhevsky2014cifar} dataset and large-scale ImageNet dataset \cite{deng2009imagenet}. The experimental results show that under the same FLOPs budget or acceleration rate, our method achieves higher accuracy than other state-of-the-art filter pruning methods. For example, with half of the FLOPs (2$\times$ acceleration) of ResNet-50 \cite{he2016deep}, we can achieve 76.1\% Top-1 accuracy on ImageNet dataset, far exceeding other filter pruning methods. Our main contributions can be summarized as follows.
\begin{itemize}
	\item  We adopt a Dagger module based on pre-trained weights to model the gates for filter pruning, such that redundant filters can be pruned more accurately to the entire dataset, avoiding the issue of joint optimization of auxiliary parameters and kernel weights.
	\item We propose to involve FLOPs as an explicit regularization to guide pruning redundant filters besides the classification performance.
	\item Our method is easy to implement, and experimental results indicate our superiority to other state-of-the-art pruning methods. 
\end{itemize}

\section{Related Work}

\begin{figure*}[t]
	\centering
	\includegraphics[width=1.0\linewidth]{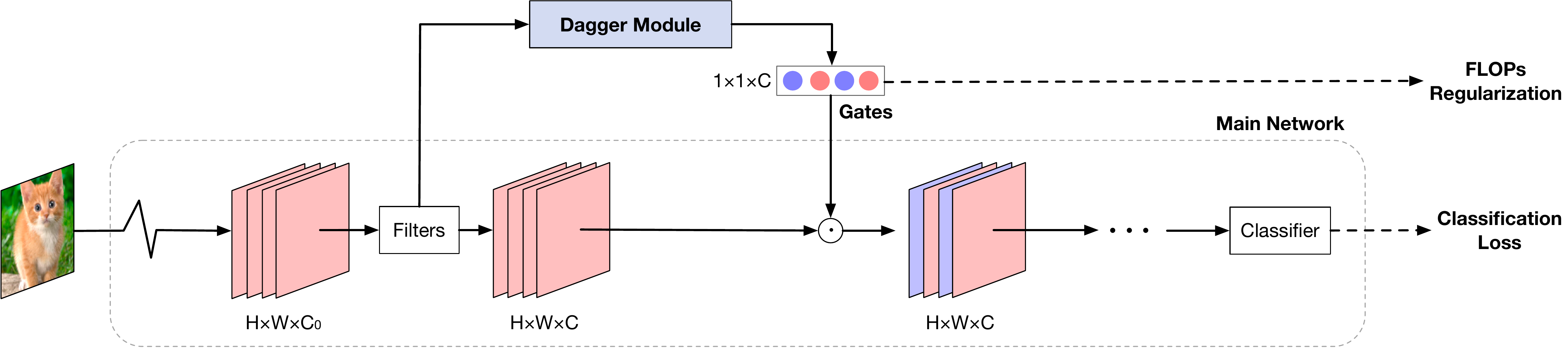}
	\caption{Overall framework of our proposed method. Dagger module is proposed that uses filter parameters to generate gates for pruning redundant filters with FLOPs regularization, which can be removed without affecting the rest of the network.}
	\label{framework}
	\vspace{-2mm}
\end{figure*}

To enable an over-parameterized convolutional neural network to be deployed in low-end computational devices, various methods have been developed to reduce the model capacity and FLOPs, such as weight pruning \cite{clip_q,learn_to_prune,dynamic_network,learn_compression}, filter pruning \cite{he2017channel,luo2017thinet,Huang_2018_ECCV,tang2020reborn,Liu2017Learning,zhuang2018discrimination,tang2019bringing}, parameter quantization \cite{ChenCompressing,WuQuantized,han2015deep}, low-rank approximation \cite{DentonExploiting,SindhwaniStructured} and so on. Essentially, weight pruning always aims to optimize the weights in an unstructured way, which makes it hard to deploy on low-end computational devices and often requires a special design to achieve acceleration. While for parameter quantization and low-rank approximation, these algorithms can be applied as a complementary method to filter pruning for further reducing the computation budget. In general, Our method can be cast into the filter pruning category.

\textbf{Filter-wise scaling factors.} Filter pruning is designed to speed up the inference of the network by pruning the redundant filters. An important task in filter pruning is to assess the redundancy of filters. Many methods leverage filter-wise auxiliary parameters to obtain the redundancy of filters, for example, Liu et al. \cite{Liu2017Learning} used the filter-wise scaling factor to prune the large network into a reduced model with comparable precision. Huang et al. \cite{Huang_2018_ECCV} proposed to use filter-wise scaling factors to indicate the redundancy of filters. Xiao et al. \cite{autoprune} proposed to prune filters through optimizing a set of trainable auxiliary parameters instead of original weights. You et al. \cite{gate_decorator} proposed to leverage filter-wise scaling factors to select redundant filters.  However, in these algorithms, filter-wise auxiliary parameters are inevitably optimized simultaneously with network parameters through multiplication, and it does not make sense to prune the redundant filters directly according to auxiliary parameters without considering the scale of filter weights. 

\textbf{Gate network for pruning.} To solve the above issue, some algorithms involve a gate network to induce pruning. For example, Zhuang et al. \cite{zhuang2018discrimination} proposed a discrimination-aware network with the additional loss to select the filters that really contribute to discriminative power. Liu et al. \cite{liu2019metapruning} proposed to leverage a meta-network to help identify the number of filters in each layer. He et al. \cite{exploring_Linear} proposed to use spectral clustering on filters to select redundant filters. Veit et al. \cite{convnets} proposed to leverage gates to define their network topology conditioned on the input batches. The AutoPruner \cite{luo2018autopruner} proposed by Luo et al.  can be regarded as a separate layer, which is attached to any convolution layer to automatically prune the filters. These methods neglect the variance of input batches to the difference in sparse structure over filters and thus lead to the variance in redundant filters.

\textbf{AutoML methods.} Since the filter pruning is generally regarded as an optimizing problem \cite{liu2018rethinking}, some works adopt AutoML~(\ie~NAS) methods \cite{darts,singlepath,atomnas,you2020greedynas,yang2020ista} to automatically search the best network structure given fixed FLOPs budget. Although AutoML-based approaches usually achieve competitive performance, they can not take full advantage of a pre-trained model and are usually computationally expensive. A typical way of AutoML methods is to optimize a wide network with various operations from a huge space as a performance evaluator, and then searching the one with the best performance from the performance evaluator for training from scratch. While our method focuses on leveraging pre-trained weights to investigate the redundancy in filters and aim to obtain a compact network for certain FLOPs budgets. 

\textbf{Contributions.} we would like to highlight the contributions and advantages of our method as illustrated in follows: (1) our method uses a Dagger module to generate gates for each filter by leveraging filter parameters as input and thus being able to generate dataset related gates, which is more suitable for pruning filters since different input batches in our method have same sparse structure over filters. While other methods (\eg~\cite{gate_decorator,autoprune,senet}) take feature maps as input, which makes their gates depend on input batches; (2) As for pruning methods that use filter-wise scale factors, (\eg~\cite{Huang_2018_ECCV,pruning_filters}),  they take only one additional parameter per filter, and lack fitting complexity to model filter redundancy more adaptively.
(3) We propose a FLOPs aware regularizer to directly pruning redundant filters from the pre-trained models with target FLOPs, while other filter pruning methods generally resort to all sorts of sparsity proxies, \eg, $\ell_{21}$ norm of filters and $\ell_{1}$ norm of filter weights or scaling factors.

\section{Modeling Redundancy of Filters with Dagger Module}

\begin{figure*}[t]
	\centering
	\includegraphics[width=\linewidth]{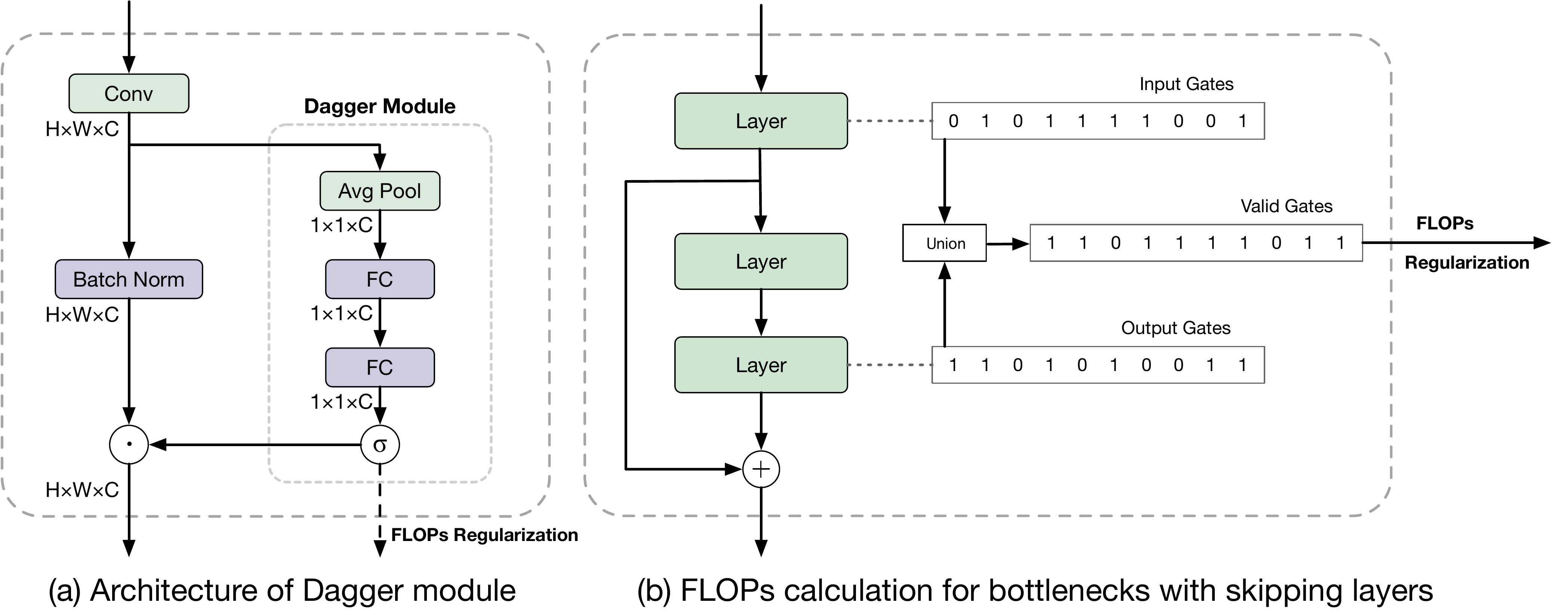}
	\caption{(a) Architecture of Dagger module. Dagger module uses convolutional filter weights to generate filter-wise gates, which are directly multiplied to convolutional feature map as final output. (b) FLOPs calculation with binary gates when dealing with skipping layers. }
	\label{fig:framework}
\end{figure*}

Filter pruning intends to identify redundant filters of the pre-trained model so that a compact pruning network with a smaller FLOPs budget can be derived. Current methods use different scaling factors, gate network or sparsity proxies to prune redundant filters, such as sparse CNN filters \cite{zhuang2018discrimination}, scaling factors of BN layers \cite{Liu2017Learning}, auxiliary factors(gates) \cite{gate_decorator,autoprune,senet} and scaling methods \cite{Huang_2018_ECCV,pruning_filters}. However, in practice, these methods generally use additional scaling factors or modeled gates with batch information as input to select the redundant filters, which makes them batch-dependent and may be harmful to locating redundant filters since different input batches may have various sparse structures over filters. In addition, there is usually a requirement that the pre-trained model should be pruned under a specified FLOPs level or accelerated by certain times. To solve the above problems, we propose to leverage the Dagger module to identify redundant filters and use pre-trained weights as input to directly obtain a network structure with certain FLOPs budget.  Denote the original network as $\N$ with pre-trained weights $\Omega^*$.

\subsection{Binary gates}
Suppose the network has $L$ layers, and feature maps for each layer $l$ are denoted as $F^l\in\mathbb{R}^{N\times n^l\times h^l\times w^l}$, where $N$ is the batch size, $n^l$ is the number of filters, and $h^l$ and $w^l$ are the spatial height and width of the feature map, respectively. In this way, to identify redundant filters based on the pre-trained model, we can allocate a binary gate $g\in\{0,1\}$ for each filter where $g=1$ means the corresponding filter should be retained, while $g=0$ is to be pruned. With the introduced gates, the feature maps can be augmented filter-wisely. Mathematically, for $l$-th layer, the augmented feature map $\tilde{F}^l$ is thus expressed as a multiplication of tensor by the scalar, \ie, 
\begin{equation}\label{merge_feature}
\tilde{F}^l_{:,i,:,:} = \g^l_i  \cdot F^l_{:,i,:,:} 
\end{equation}
where $F^l_{:,i,:,:}\in\mathbb{R}^{h^l\times w^l}$ is the $i$-th feature map of $F^l$. $\g^l_i$ corresponds to the gate of $i$-th filter \wrt~$l$-th layer. As Eq.\eqref{merge_feature}, $\g^l_i\in\{0,1\}$ controls whether the corresponding filter in $\tilde{F}^l$ is activated. When $\g^l_i$ is 0, the corresponding filters in feature map $\tilde{F}^l$ are deactivated and pruned. 

As a result, the number of retrained filters of the pruned network is directly controlled by the binary gates $\g$. Concretely, for feature map $F^l$ \wrt~$l$-th layer, its valid filter number $\tilde{n}^l$ is exactly the amount of non-zero gates, \ie,
\begin{equation}
\tilde{n}^l = \sum_{i=1}^{n^l} \I(g^l_i=1) := \norm{\g^l}_0,
\end{equation}
where $\I(\cdot)$ is an indicator function and $\g^l=\{g^l_i\} \in\{0,1\}^{n^l}$.  Specifically, for the non-pruned pre-trained model, its gates are all ones, and $\tilde{n}^l  = n^l$. Given a pre-trained model, its FLOPs are only up to the valid number of filters since the operations and spatial size $h^l\times w^l$ have been fixed. Then, the FLOPs of the pruned network is determined by the gates $\G = \{\g^l\}_{l=1}^L$, which can be written as
\begin{equation}\label{zongFlops}
\FLOPs(\G) = \sum_{l=1}^L \F^l(\g^l),
\end{equation}
where $\F^l(\cdot)$ is the FLOPs calculator \wrt~$l$-th layer. For example, if $l$-th layer has $n^l$ filters, than for a 1$\times$1 convolutional layer, its FLOPs calculator $\F^l(\g^l)$ with gate $\g^l$ will be 
\begin{equation}\label{1time1}
\F^l(\g^l) = \norm{\g^{l-1}}_0 \times h^l \times w^l \times \norm{\g^{l}}_0.
\end{equation}
As a result, we can formulate the number of filters as a mixed 0-1 binary optimization problem, 
\begin{equation}\label{formulation}
\begin{split}
\min_{\G,\Omega}~&~\L(\G,\Omega;\D_{tr},\Omega^*) \\
\st~&~\FLOPs(\G) \leq C,~\G\in\{0,1\},
\end{split}
\end{equation}
where $\Omega$ is the weights of network $\N$ with pretrained weights $\Omega^*$. $\D_{tr}$ is the training dataset. Note that the hard constraint in Eq.\eqref{formulation} can amount to a version indicated by acceleration rate $r$, \ie,
\begin{equation}\label{accelation}
\FLOPs(\G) \leq C_0/r,
\end{equation}
where $C_0$ is the overall FLOPs of the pretrained model $\N(\Omega^*)$. Then the number of filters (or gates) will be learned to minimize the training cost but under certain FLOPs budget $C$.

\subsection{Modeling gates with the Dagger module}

However, 0-1 optimization in Eq.\eqref{formulation} is an NP-hard problem. Thus we relax the original problem by considering a real-number gate in the interval $[0,1]$. Moreover, to better model a real gate, accompanied by the main network we leverage an auxiliary network named Dagger module to generate real-number gates with the help of trained weights $\Omega$. The generated filter-wise binary gates are directly applied to output feature maps with dot products for identifying redundant filters. The proposed framework is shown in Fig.  \ref{framework}.

Besides, the previous filter pruning algorithms usually take batch information (\eg~images) as input to prune the redundant filters, since different input batches may have various sparse structures over filters, the selected gates can be modeled as:
\begin{equation}
\g^l_i(\M) = \mathcal{M}(\x_i;\theta^l). 
\end{equation}
Where $\x_i$ denotes $i$-th batch information from the training dataset $\D_{tr}$. However, different batch information may lead to different redundant filters, and it is almost impossible to infer the global optimal solution of the redundancy filter of the entire dataset from the local optimal solution corresponding to the batch information.

Therefore, directly using the information related to the $\D_{tr}$ as input for the Dagger module can result in the global optimal solution about redundant filters. In detail, we use the pre-trained convolution weights $\W^l \in\mathbb{R}^{n^l\times n^{l-1}\times d_h \times d_w}$ as the input information for the Dagger module, since these weights are updated by the information of entire $\D_{tr}$ through gradient descent for several epochs. The gate $\g^l$ is supposed to be generated by taking kernel weights $\W^l$ as input, which makes the computation of the Dagger module independent from the input batches, thereby eliminating the batch variation over filters and avoiding the issue of joint optimization with kernel weights. The generation of gates can be modeled via a network (\ie, Dagger module) denoted as $\mathcal{M}$ with Dagger-weights $\theta^l$, namely,
\begin{equation}
\g^l(\M) = \mathcal{M}(\W^l;\theta^l). 
\end{equation}
The structure of our adopted Dagger module is shown in Fig. \ref{fig:framework}(a). To reduce the computation complexity of the Dagger module, we first merge the filter $\W^l$ by average pooling so that it will have the same size as the gates $\g^l$. Then the merged filters are passed through two simple fully-connected (FC) layers and further activated via a sigmoid function $\sigma(\cdot)$, \ie,
\begin{equation}
\g^l(\M) = \Sigmoid(\FC(\ReLU(\FC(\mbox{AvgPooling}(\W^l))))).
\label{Gates}
\end{equation}
Note the sigmoid function is used for mapping the gates into interval $(0,1)$. Besides, since the CNN filters may have some different magnitude of values, we also implement a normalization before these two FC layers by subtracting the mean.

By using the Dagger module, the gates can be modeled continuously, and gates approximating zero are thus reckoned to be redundant, so their corresponding filters are supposed to be pruned. Remark SE \cite{hu2018squeeze}, we do not model the gates as the squeeze-and-excitation (SE) module and Autopruner \cite{luo2018autopruner} for we can discard the Dagger module after the pruning since it is independent with input batches. However, the modules in SE and Autopruner both use feature maps as input to the auxiliary network, which will cause their gates to be batch dependent, so they can only produce sub-optimal batch related pruning results. 

\section{Pruning with FLOPs-aware Regularization}

With the gates generated by Dagger module, the original Eq.\eqref{formulation} has been relaxed into a continuous optimization problem. However, the hard constraint of FLOPs in Eq.\eqref{formulation} depends on the $\ell_0$ norm of gates $\g^l$, which is not computationally feasible for optimization. In this case, we approximate it by adopting the surrogate $\ell_1$ norm. For example, the FLOPs of Eq.\eqref{1time1} can be estimated as
\begin{equation}
\R^l(\g^l;\M) = \norm{\g^{l-1}}_1 \times h^l \times w^l \times \norm{\g^{l}}_1.
\end{equation}
And the total FLOPs in Eq.\eqref{zongFlops} can also be estimated as 
\begin{equation}\label{flops}
\R(\G;\M) = \sum_{l=1}^L \R^l(\g^l;\M).
\end{equation}
In this case, if the augmented network is initialized with all-one gates, $\R(\G;\M)$ will be an accurate estimation of FLOPs since they are equal to each other. Then minimizing $\R(\G;\M)$ will lead the gates to decrease to different values. The differences among gates reflect their different importance and sensitivity with respect to the FLOPs calculation, which amounts to the different redundancy over filters. As a result, $\R(\G;\M)$ can be regarded as a regularization for reducing the FLOPs of the pre-trained model. Moreover, since the regularization $\R(\G;\M)$ is continuous,  it enables the optimization to resort to various gradient-based optimizers, such as stochastic gradient descent (SGD). 

\subsection{pruning filters under accurate estimation}
To identify the redundant filters, the gates are supposed to also accommodate a better accuracy performance. Hence, we propose to learn them under the supervision of classification performance as well as the Dagger module and FLOPs-aware regularization. However, loss function defined in Eq.\eqref{formulation} can't be directly optimized through gradient descent. To solve this issue, we reformulated Eq.\eqref{formulation} according to lagrange multiplier \cite{lm} and leverage Eq.\eqref{flops} as the estimation of FLOPs. Then the optimization problem can be formulated as:
\begin{equation}\label{loss}
\L_{all} = \L(\G,\Omega,\M;\D_{tr},\Omega^*)  + \lambda\cdot  \R(\G;\M),
\end{equation}
where balance $\lambda>0$ is the coefficient of lagrange multiplier and will be detailed discuss in Sec V. However, the regularization is not always a good estimation due to the gap between $\ell_0$ and $\ell_1$ norm, which means that it might not be sensible to simply adopt Eq.\eqref{loss} for learning gates in an end-to-end manner. 

Inspired by the fact that $\R(\G;\M)$ is an accurate estimation of FLOPs if all gates are ones, we propose to greedily prune redundant filters. Concretely, we proceed from the pre-trained model, and all gates are initialized to be ones. In consequence, we can safely minimize the loss Eq.\eqref{loss} to learn gates since the estimation of $\R(\G;\M)$ is now exact and accurate. As a result, under the supervision and FLOPs guidance, gates with different values will be obtained. Different values indicate the current redundancy differences over corresponding filters. As a result, we greedily prune those gates with smaller values.

Nevertheless, since the gates are imposed on the feature maps, they are highly coupled in the magnitude of values. This implies that greedy pruning gates entangled with optimizing the weights may not do the trick. Therefore, we adopt an iterative update strategy, \ie, during the learning, we optimize one while fixing the other. Our proposed algorithm works in an iterative manner, as illustrated in Algorithm \ref{alg}.

\begin{algorithm}[tb]
	\caption{Data agnostic filter gating for efficient deep networks}
	\label{alg}
	\begin{algorithmic}[1]
		\REQUIRE{A well-trained model $\N$ with weights $\Omega^*$. Training dataset $\D_{tr}$. FLOPs budget $C$. All gates set $\G$.}
		\STATE initialize Dagger module $\M$ with Dagger weights $\Theta$
		\STATE one-gate set $\A = \G$, zero-gate set $\B = \emptyset$ 
		\WHILE{FLOPs$>$C}
		\STATE align gates in one-gate set $\A$
		\STATE optimize the gates with fixed weights $\Omega^*$
		\STATE get the smallest gates with ratio $r$ as $\Z$
		\STATE update $\B = \B\bigcup \Z$ and $\A = \A - \Z$
		\STATE calculate the valid FLOPs via $\A$
		\STATE fine-tune the weights with fixed all gates in $\A$ being ones
		\ENDWHILE
		\ENSURE{retained gates $\A$}
	\end{algorithmic}
\end{algorithm}

\subsection{Iterative optimization with FLOPs examination}
For a clear presentation, we refer to the original network $\N$ as \textit{main network} in contrast with the Dagger module.  As previously illustrated, we implement the iterative update for the gates (Dagger module $\mathcal{M}$) and the weights of the main network $\N$, which is presented in Algorithm \ref{alg} and elaborated as follows.

\subsubsection{Greedy pruning gates with fixed weights}
In specific, with the main network fixed, the Dagger module acts as a pruner to generate gates for the main network, which provides guidance on how to prune redundant filters in the main network. Besides, the main network supplies the gates with a classification evaluation, so that the retained gates can maintain the classification performance as much as possible. As a result, the Dagger module can be optimized by the following objective:
\begin{equation}\label{loss_M}
\L_{\M} = \L(\G,\M;\D_{tr},\Omega)  + \lambda\cdot  \R(\G;\M),
\end{equation}
where $\Omega$ is fixed compared to Eq.\eqref{loss}. Therefore, the gates can be optimized under the mutual supervision of classification loss and FLOPs-aware regularization. 

However, before we optimize the Dagger module as well as the gates, we need to fix the estimation gap of FLOPs regularization $\R(\G;\M)$. In our method, we propose to make an alignment of gates generated by the Dagger module $\M$, as line 4 of Algorithm \ref{alg}. Concretely, we first retrain the Dagger module to enable its output gates to be $0.5$, which amounts to that the second output of the FC layer in Fig.\ref{fig:framework} equals to $0$. Then we add $0.5$ to the output gates so that the values of gates are equal to $1$. The advantages of this aligning gates are two-folds. First, after alignment, the gates are all ones, thus $\R(\G;\M)$ can be an accurate estimation of FLOPs for further regularizing the redundant gates. Second, aligning gates with $0$ prior to the sigmoid activation corresponds to its maximum slopes, which in a way enhances the impact of regularization $\R(\G;\M)$ for optimizing gates. 

After the alignment, we can safely optimize Eq.\eqref{loss_M} to obtain gates. However, the estimation gap may be enlarged by optimization. In this way, we propose to prune redundant filters, \ie, greedy pruning some gates for multiple times until the retained gates satisfy the FLOPs budget. Generally, the closer a gate is to zero, the smaller its contribution to the network. Besides, since the gates are all ones after the alignment, we optimize Eq.\eqref{loss_M} for some steps (line 5 of Algorithm \ref{alg}), and then prune the gates with smaller gates. Usually, we can set a pruning ratio of $r$ (\eg, 0.6\%) to control the number of pruned gates for each update  (line 6 of Algorithm \ref{alg}). In addition, to meet a hard FLOPs constraint, we can simply implement a FLOPs examination after pruning gates each time (line 8 of Algorithm \ref{alg}). If the currently retained gates satisfy the FLOPs budget, redundant filters are expected to be learned well. As a result of using the greedy algorithm, based on pretrained weights, our algorithm can use only a small number of input batches to prune redundant filters, so as to achieve the purpose of rapid filter pruning.

\subsubsection{Fine-tuning weights with fixed gates}
After some gates are pruned (set as fixed zero), we need to fine-tune the weights of the main network with fixed gates. However, after the greedy pruning, the values of those retained gates are no longer ones but in $(0,1)$. If we implement fine-tuning weights based on them, it will further worsen the coupled issue since the weights are trained from biased gates. So we propose to set all retrained gates to ones (line 9 of Algorithm \ref{alg}), and then implement fine-tuning afterward. The objective goes as:
\begin{equation}\label{loss_O}
\L_{\Omega} = \L(\Omega;\D_{tr},\G),
\end{equation}
which compensates for the lost information in retained filters due to filter pruning.

\subsection{Dealing with skipping layers}
To construct the FLOPs-aware regularization $\R(\G;\M)$, the FLOPs needs to be calculated. For a regular layer (\eg, 1$\times$1 convolution in Eq.\eqref{1time1}), the number of filters within different layers is independent, thus we can use the $\ell_0$ norm of gates for each layer to represent the number of filters, and calculate the FLOPs in a simple form. However, for those bottlenecks with skipping layers, there is a structural constraint that the input and output of the bottleneck should have the same number of filters, such as the ResNet \cite{he2016deep} and MobileNetV2 \cite{mobilenetv2}. 

For a bottleneck with skipping layers as Fig. \ref{fig:framework}(b), each layer will have its own gates $\g^l$. Denote the gates of input and output as $\g^{in}$ and $\g^{out}$, respectively. For a pretrained model, the size of $\g^{in}$ and $\g^{out}$ is the same. Then, to calculate the FLOPs of this bottleneck, the valid gates of $\g^{in}$ and $\g^{out}$ should be their union as Fig.  \ref{fig:framework}(b) shows, \ie,
\begin{equation}
\g = \g^{in} \vee \g^{out} = 1 - (1-\g^{in})\cdot (1-\g^{out}),
\label{skip_line}
\end{equation}
where $\vee$ is the union operation. Then  based on the valid gates, the FLOPs can be calculated as regular layers.

\section{Experimental Results} 
In this section, we implement extensive experiments on benchmark CIFAR-10 and ImageNet datasets to validate the superiority of our proposed method. Besides, we also conduct ablation studies to further investigate how our method contributes to methods of filter pruning.

\subsection{Configuration and settings} 
\textbf{Comparison methods}. In order to compare pruning performance, we select several state-of-the-art filter pruning methods, AutoPruner \cite{luo2018autopruner}, LEGR \cite{legr}, SFP \cite{sfp}, FPGM \cite{fpgm}, DCP \cite{zhuang2018discrimination}, ThiNet   \cite{luo2017thinet}, CP \cite{he2017channel}, Slimming \cite{Liu2017Learning} and PFS \cite{pfs}. Besides, since our method aims to identify redundant filters, we also cover two vanilla baselines. The first one is \textit{Uniform}, \ie, shrinking the width of a network by the fixed rate to meet the requirement of FLOPs budget. The second one is a variant of the random set of filters within each layer, denoted as \textit{Random}. Concretely, we randomly adjust the number of filters within Uniform in a certain range to meet the FLOPs budget. The Random method is implemented for 10 times, and we report the average performance.

\textbf{Training.} Based on a pre-trained model, we prune redundant filters until the FLOPs budget is satisfied. Specifically, we optimize the Dagger module and main network for $400 (100)$ iterations with a batch size of $320 (64)$ for ImageNet (CIFAR-10) dataset before pruning the gates in each update in Algorithm \ref{alg}. The pruning rate per update (line 6 in Algorithm \ref{alg}) is set to 0.6\% and the balance parameter $\lambda$ is set to 8 for all networks. We use SGD optimizer with momentum 0.9 and nesterov acceleration. The weight decay is set to $0.0001$. Besides, the learning rate is annealed with a cosine strategy from initial value 0.001 for Dagger module (main network). Once the FLOPs budget is achieved, we will finetune the pruned weights with the learning rate initialized to 0.01. For the CIFAR-10 dataset, we finetune 100 epochs and the learning rate is divided by 10 at {75-th, 112-th} epoch. For the ImageNet dataset, we use the cosine learning rate to finetune the network for 60 epochs. All experiments are implemented with PyTorch \cite{paszke2017pytorch} on NVIDIA 1080 Ti GPUs.

\begin{table*}[t]
	\centering
	\caption{Performance comparsion of MobileNetV2 and VGGNet on CIFAR-10.}
	\label{results:cifar}
	\renewcommand\tabcolsep{0.8pt} 
	\resizebox{\textwidth}{!}
	{\begin{tabular}{c|c|ccc|c|c|ccc}
			\hline
			\multicolumn{5}{c|}{MobileNetV2}&\multicolumn{5}{c}{VGGNet} \\ \hline
			Groups&Methods&FLOPs&Params&Acc&Groups&Methods&FLOPs&Params&Acc  \\ \hline
			\multirow{4}*{200M} & DCP \cite{dcp} & 218M & - & 94.69\% & \multirow{8}*{200M} & DCP \cite{dcp} & 199M & 10.4M & 94.16\% \\
			& Uniform & 207M & 1.5M & 94.57\% & & Slimming \cite{Liu2017Learning} & 199M & 10.4M & 93.80\% \\
			& Random & 207M & - & 94.20\% & & SSS \cite{Huang_2018_ECCV} & 199M & 5.0M & 93.63\% \\
			& \textbf{Dagger} & 207M & 1.9M & \textbf{94.91\%} & & PFS \cite{pfs} & 199M & - &  93.71\% \\ \cline{1-5}
			\multirow{4}*{148M} & MuffNet \cite{muffNet} & 175M & - & 94.71\% & & VCN \cite{vcn} & 190M & 3.92M & 93.18\% \\
			& Uniform & 148M & 1.1M & 94.32\% & & Uniform & 199M & 10.0M & 93.45\%\\
			& Random & 148M & - & 93.85\% & & Random & 199M & - & 93.02\% \\
			& \textbf{Dagger} & 148M & 1.7M & \textbf{94.83\%} & & \textbf{Dagger} & 199M & 6.0M & \textbf{94.25\%} \\ \hline
			\multirow{5}*{88M} & AutoSlim \cite{autoslim} & 88M & 1.5M & 93.20\% & \multirow{5}*{119M} & AOFP \cite{aofp} & 124M & - & 93.84\%  \\
			& Uniform & 88M & 0.6M & 94.32\% & & CGNets \cite{CGNets} & 117M & - & 92.88\%\\
			& Random & 88M & - & 93.85\% & & Uniform & 119M & 6.1M & 93.03\% \\
			& \textbf{Dagger} & 88M & 1.1M & \textbf{94.49\%} & & Random & 119M & - & 92.22\% \\
			& AutoSlim \cite{autoslim} & 59M & 0.7M & 93.00\% & & \textbf{Dagger} & 119M & 2.7M & \textbf{93.91\%} \\ \hline
			
	\end{tabular}}	
	\vspace{-2mm}
\end{table*}

\subsection{Experiments on CIFAR-10 dataset}
\textbf{Dataset and networks.} The CIFAR-10 dataset includes 60,000 RGB images of 32$\times$32 sizes from 10 exclusive categories. The dataset includes 50,000 images for training and 10,000 images for testing. We conduct filter pruning on the benchmark VGGNet-19 \cite{simonyan2014very} and the compact MobileNetV2 \cite{mobilenetv2}. Concretely, VGGNet-19 has 20M parameters and 399M FLOPs with an error rate of 6.01\%. In contrast, MobileNetV2 only has 2.2M parameters and 297M FLOPs but with an error rate of 5.53\%. The results are reported in Table \ref{results:cifar}.

\textbf{Results.} As shown in Table \ref{results:cifar}, our method achieves the best accuracy \wrt~different FLOPs on MobileNetV2 and VGGNet. In detail, for VGGNet, our pruned 50\% FLOPs VGGNet outperforms the state-of-the-art DCP, Slimming, and PFS by 0.26\%, 0.45\%, and 0.54\%, respectively, and even surpass the pretrained model by 0.26\%, which means our method can efficiently prune redundant filters, thereby improving performance. In addition, compared with the two baselines Uniform and Random, our pruned VGGNet-19 can improve the accuracy by more than 0.80\%. Different from VGGNet-19, MobileNetV2 is more compact and has many skipping layers. As shown in Table \ref{results:cifar}, the performance of our 207M MobileNetV2 can outperform DCP and the pretrained model by 0.44\% and 0.22\%, respectively. Besides, in the case of a tiny budget (\ie~88M FLOPs), our pruned MobileNetV2 can still achieve an accuracy of 94.49\%, and it is 1.29\% higher than AutoSlim, which proves that our method can achieve promising results even with small budgets.

\begin{table*}[t]
	\centering
	\caption{Performance comparison of pruned ResNet-50 on ImageNet dataset with $\sim$2.0G FLOPs budget.}
	\label{ResNet50}
	\renewcommand\tabcolsep{3.0pt} 
	\resizebox{0.77\textwidth}{!}
	{\begin{tabular}{c||cccc}
			\hline
			Methods&FLOPs&Params&Top-1 ACC&Top-5 ACC  \\ \hline
			SFP \cite{sfp} & 2.4G & - & 74.6\% & 92.1\% \\
			FPGM \cite{fpgm} & 2.4G & - & 75.6\% & 92.6\% \\
			LEGR \cite{legr} & 2.4G & - & 75.7\% & 92.7\% \\
			AutoPruner \cite{luo2018autopruner} & 2.0G & - & 74.8\% & 92.2\% \\
			MetaPruning \cite{liu2019metapruning} & 2.0G & - & 75.4\% & - \\
			Uniform & 2.0G & 10.2M & 74.1\% & 90.6\% \\
			Random & 2.0G & - & 73.2\% & 90.4\% \\ 
			\textbf{Dagger} & 2.0G & 11.7M & 76.1\% & \textbf{92.8\%} \\ \hline
			
	\end{tabular}}	
	\vspace{-2mm}
\end{table*}

\begin{table*}[t]
	\centering
	\caption{Performance Comparison of pruned MobileNetV2 on ImageNet dataset with two FLOPs budget.}
	\label{MobileNet_ImgNet}
	\renewcommand\tabcolsep{5.0pt} 
	\resizebox{0.90\textwidth}{!}
	{\begin{tabular}{c|c||cccc}
			\hline
			Groups&Methods&FLOPs&Params&Top-1 ACC&Top-5 ACC  \\ \hline
			\multirow{5}*{140M} & MetaPruning \cite{liu2019metapruning} & 140M & - & 68.2\% & - \\
			& GS \cite{gs} & 137M & 2.0M & 68.8\% & - \\
			& Uniform & 140M & 2.7M & 67.6\% & 88.2\% \\
			& Random & 140M & - & 67.1\% & 87.9\% \\
			& \textbf{Dagger} & 140M & 2.84M & 69.5\% & \textbf{88.8\%} \\ \hline
			\multirow{5}*{106M} & MetaPruning \cite{liu2019metapruning} & 105M & - & 65.0\% & - \\
			& GS \cite{gs} & 106M & 1.9M & 66.9\% & - \\
			& Uniform & 106M & 1.5M & 64.1\% & 84.2\% \\
			& Random & 106M & - & 63.5\% & 84.0\% \\
			& \textbf{Dagger} & 106M & 2.46M & 67.2\% & \textbf{86.8\%} \\ \hline
			
	\end{tabular}}	
	\vspace{-2mm}
\end{table*}

\subsection{Experiments on ImageNet dataset}
\textbf{Dataset.} The ImageNet (ILSVRC-12) dataset consists of 1.28 million training images and 50k validation images from 1000 categories. In specific, we report the accuracy of the validation dataset as \cite{zhuang2018discrimination,Liu2017Learning}. Then we implement pruning on two benchmark networks, \ie, ResNet-50 and MobileNetV2. The pretrained models refer to those released by Pytorch. \footnote{\url{https://pytorch.org/docs/stable/torchvision/models.html}}

\textbf{Results of ResNet-50.} The pretrained ResNet-50 has 25.5M parameters and 4.1G FLOPs with 76.6\% Top-1 accuracy. As shown in Table \ref{ResNet50}, our algorithm outperforms the SFP \cite{sfp}, FPGM \cite{fpgm} and LEGR \cite{legr} by 1.5\%, 0.5\% and 0.4\%, respectively, while our pruned ResNet-50 has even smaller FLOPs (by 0.4G). Besides, from comparsion with AutoPruner \cite{luo2018autopruner} and MetaPruning \cite{liu2019metapruning}, our pruned network also achieves the best performance with 1.3\% and 0.7\% improvement on Top-1 accuracy. 

\textbf{Results of MobileNetV2.} The pretrained MobileNetV2 has 3.5M parameters and 300M FLOPs with 68.2\% Top-1 accuracy. We prune the network under two different FLOPs budgets (140M and 106M). As shown in Table \ref{MobileNet_ImgNet}, by pruning the MobileNetV2 to 140M FLOPs, our pruned MobileNetV2 outperforms the pretrained MobileNetV2 by 1.3\%. Our method also leads to 2.4\% and 1.9\% increase in Top-1 accuracy compared with the baseline of Random and Uniform for FLOPs 140M (106M). Moreover, with the same FLOPs budget, our method can actually surpass MetaPruning by a large margin of 1.3\% for FLOPs 140M (106M).

\begin{figure*}[t]
	\centering
	\subfigure
	{\includegraphics[width=0.48\columnwidth]{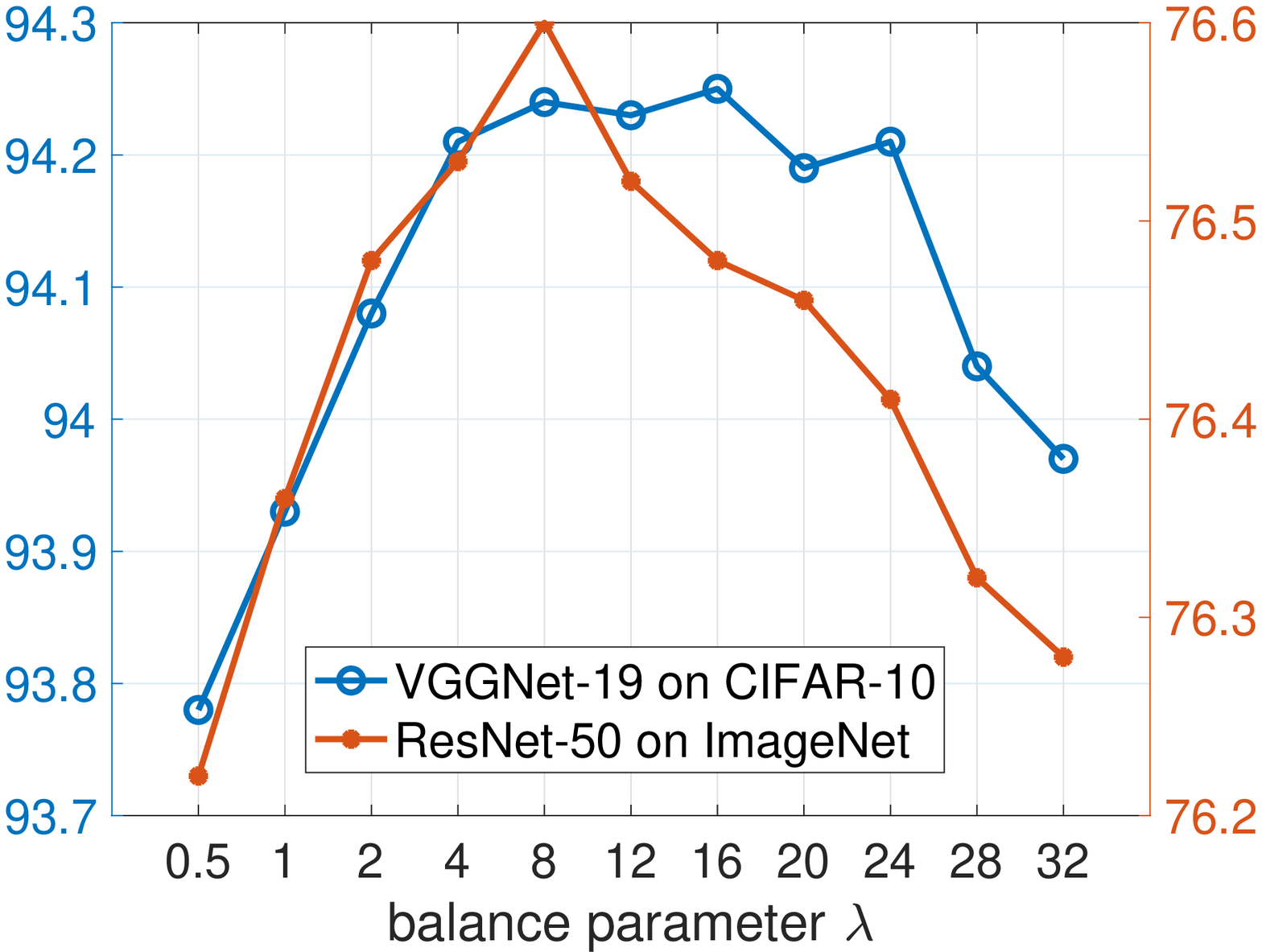}}
	~~\subfigure
	{\includegraphics[width=0.48\columnwidth]{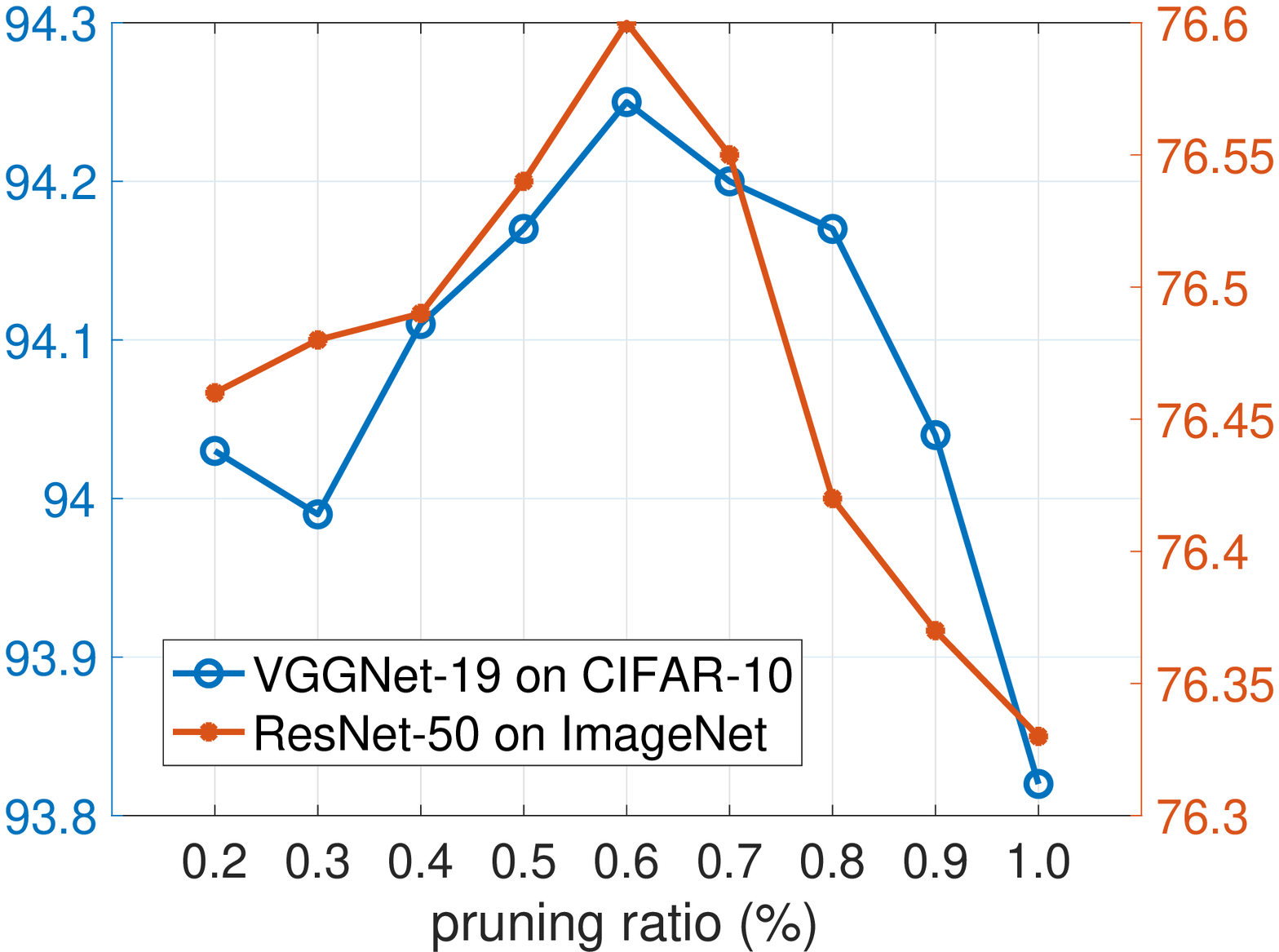}}
	\caption{Ablation studies. Classification performance of the pruned networks under different balance parameter $\lambda$ in Eq.\eqref{loss_M} (Left) and pruning ratios (Right). Note that the \blue{blue} lines refer to the Top-1 accuracy (\%) of VGGNet-19 on CIFAR-10 dataset while the \red{red} lines are for the Top-1 accuracy (\%) of ResNet-50 on ImageNet dataset.}
	\label{fig:ablation}
	\vspace{-2mm}
\end{figure*}

\subsection{Ablation studies}

\subsubsection{Effect of balance parameter $\lambda$}

According to the lagrange multiplier, the value of $\lambda$ is achieved when the derivative of the variable in Eq.\eqref{loss_M} is 0, which means the loss has reached the minimum point. And the relationship between loss and gates cannot be expressed explicitly, we can't get the exact value of $\lambda$ through calculation. However, we can experimentally find the value of $\lambda$ that makes loss or accuracy achieves minimum point. In detail, we implement pruning with different trade-off parameters $\lambda\in\{0.5, 1, 2, 4, 8, 16, 20, 24, 28, 32\}$. When the trade-off parameter becomes larger, the impact of the classification loss will become smaller, and the network will pay more attention to reducing FLOPs, resulting in a rapid increase in the classification loss with more iterations involved. On the other hand, if $\lambda$ is too small, the network is likely to be randomly pruned, leading to poor retraining results. The accuracy of each pruned network with different $\lambda$ is shown in Fig.\ref{fig:ablation}. Empirically, we set $\lambda$ to 8 for all networks.

\subsubsection{Effect of greedy pruning }
In our method, the main network is updated after each pruning process. In this way, the pruning ratio $r$ in line 6 of Algorithm \ref{alg} not only controls the pruning speed of each update but also changes the strategy of pruning. Since we subtract the mean of filters in the Dagger module when the pruning ratio is chosen to be a larger value, the gates will be pruned more evenly and vice versa. In order to investigate the effect of pruning ratio, we prune VGGNet-19 on CIFAR-10 dataset and ResNet-50 on ImageNet dataset with different pruning ratios, \ie, $\{ 0.2\%,0.3\%,0.4\%, ..., 0.9\%,1.0\%\}$. As shown in Fig. \ref{fig:ablation}, the pruning ratio favors a medium value since a large value tends to uniformly prune gates over all layers while a smaller value will greedily prune a certain layer. We find 0.6\% is empirically a good option.

\begin{figure*}[t]
	\centering
	\includegraphics[width=\linewidth]{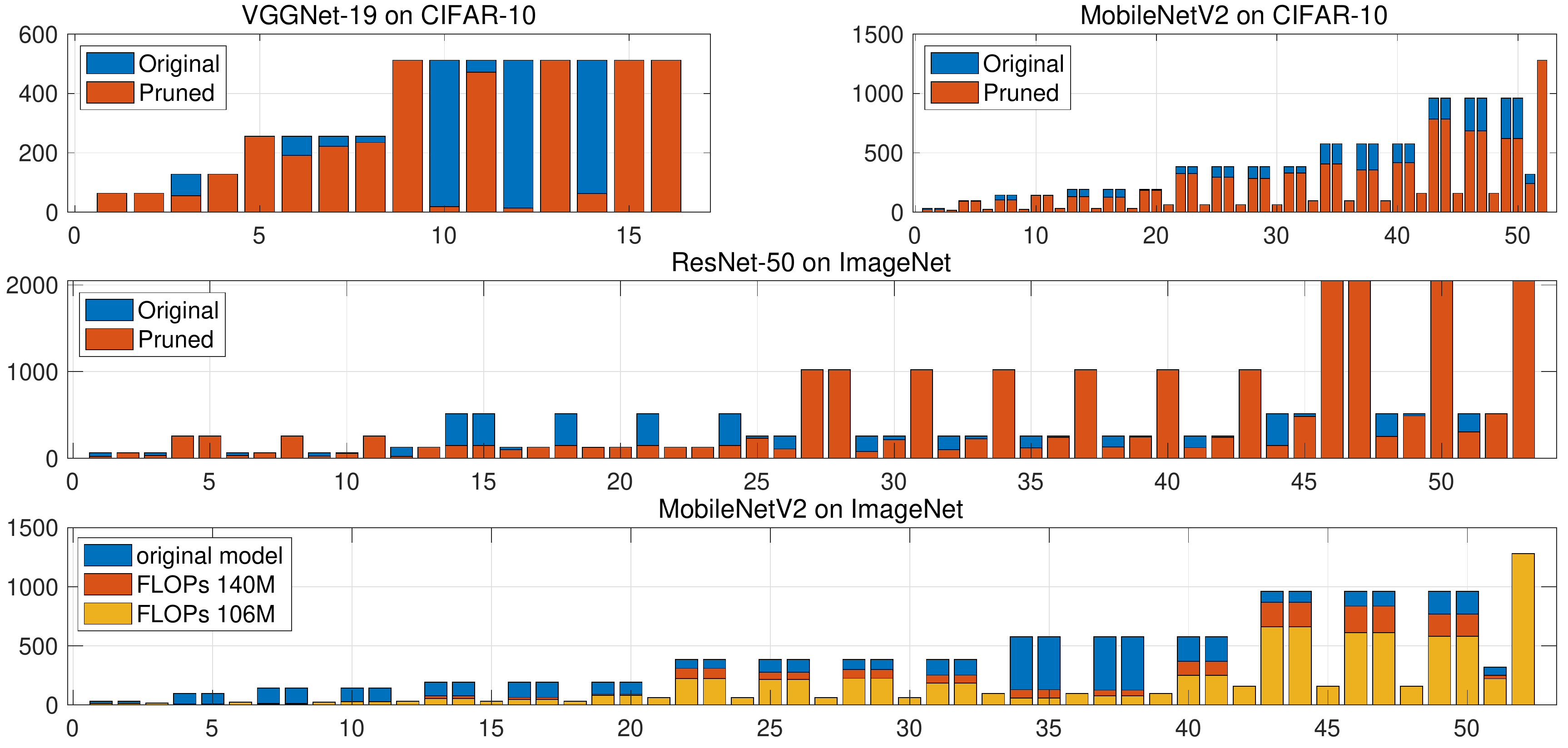}
	\caption{Visualization of the learned number of filters \wrt different networks and datasets.}
	\label{fig:mobilenet}
	\vspace{-3mm}
\end{figure*}

\subsubsection{Visualization of learned number of filters}
Based on the CIFAR-10 and ImageNet dataset, Fig. \ref{fig:mobilenet} shows our pruned results for VGGNet-19, MobileNetV2 and ResNet-50. For two networks with skipping layers (ResNet-50 and MobileNetV2), the pruning is smoothly distributed to all layers. However, MobileNetV2 pays more attention to pruning layers that contain skipping layers, while ResNet-50 does not. The network structure of VGGNet-19 does not contain any skipping layers, so its pruning results are more uneven than other networks with skipping layers. In addition, the last layers of the above three networks are well preserved after pruning, which may result from they contribute more to the final classification. 

To analyze the pruning process of the same network when different FLOPs budgets are given, as shown in Fig. \ref{fig:mobilenet}, we prune MobileNetV2 on ImageNet dataset with 140M and 106M FLOPs budgets, respectively. When MobileNetV2 is pruned from 300M to 140M, the pruning process is mainly concentrated on the non-skipping layers and those layers near the front of the network. However, when the FLOPs budget is set to 106M, the number of filters at the end of the network begins to decrease, implying that the layers in front of the network are easier to be pruned than the end layers.

\begin{table}[t]
	\centering
	\caption{Classification error (\%) of pruned networks on CIFAR-10 dataset with 2x acceleration rate \wrt~different checkpoints of pretrained models.}
	\label{pretrain_test}
	\renewcommand\tabcolsep{5.0pt} 
	\renewcommand\arraystretch{0.7} 
	\resizebox{0.78\textwidth}{!}
	{\begin{tabular}{c|c|c||c}
		\hline
		Model & Pretrain Epochs & Pretrain ACC & Finetune ACC\\
		\hline
		\multirow{7}*{MobileNetV2} & 10 & 87.89\%&94.51\% \\
		& 40 & 89.86\%&94.63\% \\
		& 70 & 90.74\%&94.74\% \\
		& 100 & 91.89\%&94.77\% \\
		& 150 & 93.40\%&94.78\% \\ 
		& 300 & 94.47\%&94.83\% \\ \hline
		\multirow{7}*{VGGNet} & 10 & 85.25\%&94.07\% \\
		& 40 & 89.51\%&94.13\% \\
		& 70 & 90.57\%&94.17\% \\
		& 100 & 91.55\%&94.22\% \\
		& 150 & 92.76\%&94.21\% \\ 
		& 300 & 93.99\%&94.25\% \\ \hline
	\end{tabular}}
	\vspace{-3mm}
\end{table}

\begin{figure}[t]
	\centering
	\includegraphics[width=0.66\linewidth]{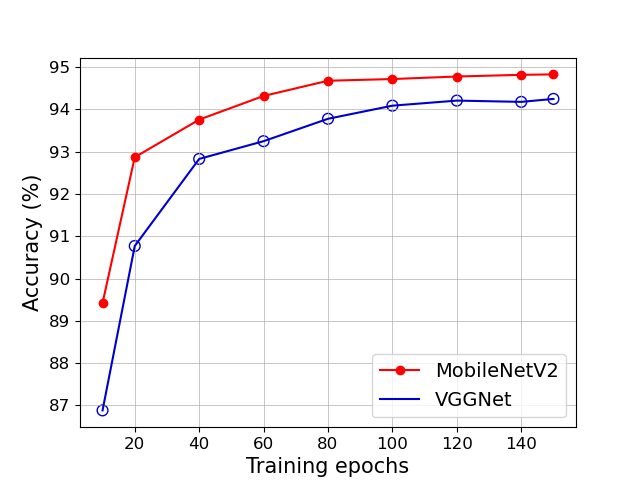}
	\caption{Finetuning epochs of 2 $\times$ acceleration of MobileNetV2 and VGGNet on CIFAR-10 dataset.} 
	\label{finetune_ep}
	\vspace{-2mm}
\end{figure}

\subsubsection{Effect of quality of pretrained models}\label{diff_pretrain}

To examine the effect on learned filter numbers with different quality of pretrained models, we use different checkpoints of the pretrained  MobileNetV2 (VGGNet) on the CIFAR-10 dataset, which have different classification errors. Then we implement our method Dagger based on these pretrained models, and the results are shown in Table \ref{pretrain_test}. In detail, we use pretrained models \wrt~different pre-train epochs as the pretrained model to implement our method Dagger, the accuracy of pretrained models are referred to as "Pretrain Acc". After pruning with  2 $\times$ acceleration, we finetune the retained weights with 100 epochs and report as "Finetune Acc". It can be seen that with the improvement of classification performance of pretrained models, our pruned networks get improved as well. Moreover, the improvement tends to be steady if the quality of the pretrained model is not too bad. For example, when the pretrained epochs of MobileNetV2 (VGGNet) are changed from 300 to 70 epochs, its pretrained accuracy degrades for 3.73\% (3.42\%), while our finetune accuracy of the pruned results only have 0.09\% (0.07\%) performance gap. This implies that our method shows small sensitivity towards the quality of the pretrained model; the pretrained models do not necessarily need to be state-of-the-art ones but not-too-bad ones if we expect a good pruned network.

\subsubsection{Effect of Dagger in retraining epochs}
To examine the effect of the finetuning epochs in our method Dagger. We finetune the pruned results of 50\% FLOPs MobileNetV2 and VGGNet in CIFAR-10 dataset \wrt~different epochs. In detail, we inherit the weights of the retaining filters after pruning and adopt the same training strategy as illustrated before. Specifically, for VGGNet, the learning rate is initialized to 0.1 and divided by 10 at 50\% and 75\% of the total epochs. As shown in Fig. \ref{finetune_ep}, the test accuracy of the pruned models at the initial remains relatively low, which means that the lost information of the pruned filters has a certain effect on the overall performance. However, as the finetuning epochs increases, the accuracy of the pruned models improves sharply, reaching the highest accuracy with about 100 epochs, proving the effectiveness of our method.

\subsubsection{Efficiency of Dagger in pruning filters}
To investigate the efficiency of our method in pruning filters, we report the time cost on pruning \wrt~different pruning ratios in Table \ref{model_complexity}. All experiments are implemented with 8 NVIDIA 1080 Ti GPUs.

\begin{table}[t]
	\centering
	\caption{Efficiency of pruned ResNet-50 and MobileNetV2 on ImageNet Dataset \wrt~different pruning ratios.}
	\label{model_complexity}
	\renewcommand\tabcolsep{5.0pt} 
	\renewcommand\arraystretch{0.80} 
	\resizebox{0.77\textwidth}{!}
	{\begin{tabular}{c|c|ccc}
		\hline
		Model & Pruning ratios & Params & FLOPs & Time cost(h)\\ \hline
		\multirow{3}*{ResNet-50} & 30\%& 16.5M& 2.9G& 1.5$\times$8\\
		& 50\%& 11.7M& 2.0G& 2.5$\times$8\\
		& 70\%& 9.2M& 1.2G& 3.4$\times$8\\  \hline
		\multirow{3}*{MobileNetV2} & 30\%& 3.19M& 210M& 1.0$\times$8\\
		& 50\%& 2.96M & 150M& 1.4$\times$8\\
		& 70\%& 2.54M& 90M&1.8$\times$8\\ \hline
	\end{tabular}}
	\vspace{-2mm}
\end{table}

As shown in Table \ref{model_complexity}, our method can quickly get the desired model size based on the pretrained model. We optimize the Dagger module and main network for 400 (100) iterations with the batch size of 320 (64) and pruning ratio 0.6\% for ImageNet (CIFAR-10) dataset in each update. Therefore, taking pruning models to 50\% FLOPs as an example, we only need to go through about 8 (10) epochs for ImageNet (CIFAR-10) dataset.
\begin{figure*}[t]
	\centering
	\includegraphics[width=0.85\linewidth]{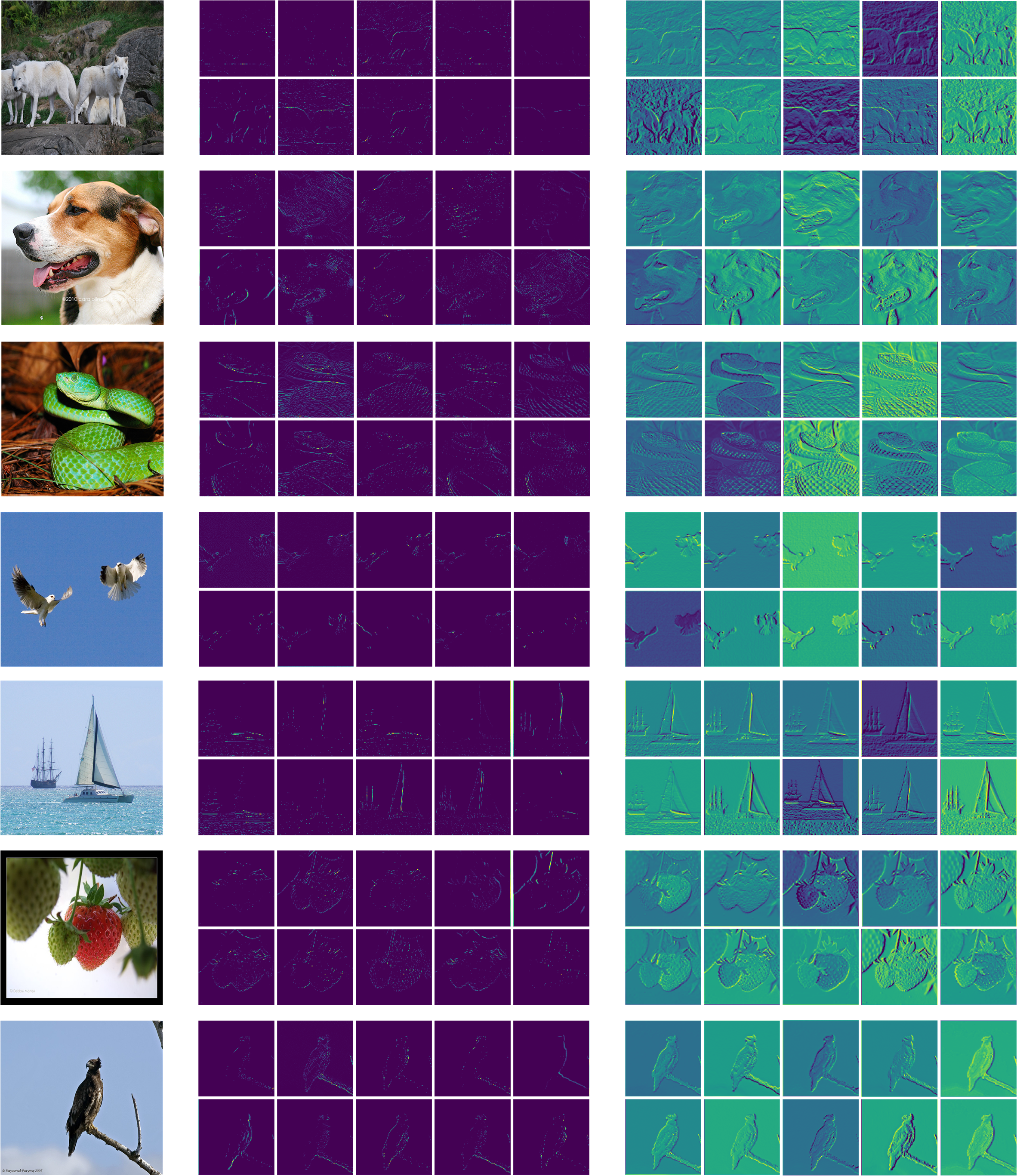}
	\caption{Visualization of feature maps \wrt~pruned (middle column) and retrained (right column) filters in second bottleneck of MobileNetV2 on ImageNet dataset.}
	\label{Feature_map}
	\vspace{-2mm}
\end{figure*}

\subsection{Visualization of feature maps}
\label{Visualization}
To intuitively check the learned gates by our method Dagger, we visualize the feature maps \wrt~different filters with zero gates (pruned) and one gate (retained) in Fig. \ref{Feature_map}. All the feature maps are from the first convolution of the second bottleneck in MobileNetV2 based on the ImageNet dataset. As shown in Fig. \ref{Feature_map}, the feature maps of retrained filters (one gate) are more visually informative than that of pruned ones (zero gates). Besides, the pruned filters usually contain more background information, \eg, the bird in the fourth line of Fig. \ref{Feature_map}. In contrast, our retained filters have a lot of information about the target and suppress background instead, such as snakes and dogs in the second and third rows of Fig. \ref{Feature_map}. 


\section{Conclusion}

In this paper, we propose to leverage Dagger module with pre-trained weights as input and involve FLOPs as an explicit regularization to guide the process of redundant filters pruning. Concretely, we assign a binary gate for each filter to indicate whether the filter should be retrained or pruned. The binary gates can be well modeled by a Dagger module with the help of CNN filters and use filter parameters as input, which helps to generate dataset related gates. Then based on aligning gates, we can have an accurate estimation of FLOPs, and it can also guide the redundant filters learning besides the classification performance. We prune the redundant filters from the pre-trained model through the greedy algorithm until the effective FLOPs of the current pruned network matches with the pre-set budget. Extensive experiments on benchmark CIFAR-10 dataset and large-scale ImageNet dataset show the superiority of our proposed method over other state-of-the-art filter pruning methods.

\bibliography{reference_ChannelPruning}

\end{document}